\documentclass{article}
\usepackage{ijcai16}
\usepackage{clrscode3e}
\usepackage{amsthm}
\theoremstyle{plain}

\theoremstyle{definition}
\newtheorem{defn}{Definition}
\theoremstyle{plain}

\usepackage{amssymb}
\usepackage{amsmath}
\usepackage{pdfcomment}
\usepackage[normalem]{ulem}

\usepackage{times}

\usepackage{enumitem}

\title{Learning to Rank for Synthesizing Planning Heuristics}
\author{Caelan Reed Garrett,  Leslie Pack Kaelbling, Tom\'as Lozano-P\'erez \\ 
MIT CSAIL\\
Cambridge, MA 02139 USA\\
\{caelan, lpk, tlp\}@csail.mit.edu}

\begin{document}

\maketitle

\begin{abstract}
We investigate learning heuristics for domain-specific planning. Prior work framed learning a heuristic as an ordinary regression problem. However, in a greedy best-first search, the {\it ordering} of states induced by a heuristic is more indicative of the resulting planner's performance than mean squared error. Thus, we instead frame learning a heuristic as a learning to rank problem which we solve using a RankSVM formulation. Additionally, we introduce new methods for computing features that capture temporal interactions in an approximate plan. Our experiments on recent International Planning Competition problems show that the RankSVM learned heuristics outperform both the original heuristics and heuristics learned through ordinary regression.

\end{abstract}

\section{Introduction}

Forward state-space greedy heuristic search is a powerful technique
that can solve large planning problems. However, its
success is strongly dependent on the quality of its heuristic. Many
domain-independent heuristics estimate the distance to the goal by
quickly solving easier, approximated planning
problems~\cite{HoffmannN01,helmert2006fast,helmert2008unifying}. While
domain-independent heuristics have enabled planners to solve a much
larger class of problems, there is a large amount of room to improve
their estimates.  In particular, the effectiveness of many
domain-independent heuristics varies across domains, with poor
performance occurring when the approximations in the heuristic 
discard a large amount of information about the problem.  

Previous work has attempted to overcome the limitations of these approximations by learning a domain-specific heuristic correction~\cite{yoon2006learning,yoon2008learning}. Yoon et al. formulated learning a correction for the FastForward (FF) heuristic~\cite{HoffmannN01} as a regression problem and solved it using ordinary least-squares regression. While the resulting planner is no longer domain-independent,
the learning process is domain independent, and the learned heuristic
is more effective than the standard FF heuristic. 

In this paper, we improve on these results by framing the learning
problem as a learning to rank problem instead of an ordinary
regression problem. This is motivated by the insight that, in a greedy
search, the ranking induced by a heuristic, rather than its numerical
values, governs the success of the planning.  By optimizing for the
ranking directly, our RankSVM learner is able to produce a heuristic
that outperforms heuristics learned through least-squares regression.

Additionally, we introduce new methods for constructing features for
heuristic learners. Like Yoon et al., we derive our features from an
existing domain-independent
heuristic~\cite{yoon2006learning,yoon2008learning}. However, our
features focus on the ordering and interaction between actions in
approximate plans. Thus, they can be based on any existing heuristic that implicitly constructs an approximate plan, such as the context-enhanced additive (CEA) heuristic~\cite{helmert2008unifying}. These features can be easily constructed and
still encode a substantial amount of information for heuristic learners.

In our experiments, we evaluate the performance of the different
configurations of our learners on several of the International
Planning Competition learning track problems~\cite{ipc2014}.  We find
that the learned heuristics using the RankSVM approach allow more
problems to be solved successfully than using the popular FF and CEA
heuristics alone.  Additionally, they significantly surpass the
performance of heuristics learned through ordinary regression.

\section{Related Work}

Prior work in learning for planning spans many types of
domain-specific planning knowledge~\cite{jimenez2012review}; our focus
in this paper is on learning heuristic functions.


Yoon et al. were the first to improve on a heuristic function using machine learning~\cite{yoon2006learning,yoon2008learning}. They centered their learning on improving the FF Heuristic~\cite{HoffmannN01}, using ordinary least-squares regression to learn the difference between the actual distance-to-go and the estimate given by the FF heuristic. Their key contribution was deriving features using the relaxed plan that FF produces when computing its estimate. Specifically, they used taxonomic syntax to identify unordered sets of actions and predicates on the relaxed plan that shared common object arguments. Because there are an exponential number of possible subsets of actions and predicates, they iteratively introduced a taxonomic expression that identifies a subset greedily based on which subset will give the largest decrease in mean squared error. This process resulted in an average of about 20 features per domain~\cite{xu2009learning}. In contrast, our features encode ordering information about the plan and can be successfully applied without any taxonomic syntax or iterative feature selection.

Xu et al. built on the work of Yoon et al. and incorporated ideas from
structural prediction~\cite{xu2007discriminative,xu2009learning}. They
adapted the learning-as-search optimization framework to the context
of beam search. They learn a discriminative model to rank the top $b$
successors per state to include in the beam search. 
In subsequent work, they used RankBoost to more reliably rank
successors by bootstrapping the predictions of action-selection
rules~\cite{xu2010iterative}. 
Although we also use a ranking approach, we use ranking as a loss
function to train a heuristic from the position of states along
a trajectory, resulting in a global heuristic that can be directly
applied to greedy best-first search. 

Arfaee et al. learned heuristics by iteratively improving on prior heuristics for solving combinatorial search problems~\cite{arfaee2011learning}. They used neural networks and user defined features. Finally, Virseda et al. learned combinations of existing heuristics values that would most accurately predict the cost-to-go~\cite{us2013learning}. However, this strategy does not use features derived from the structure of the heuristics themselves.

Wilt et al. investigated greedy heuristic search performance in several combinatorial search domains~\cite{wilt2012does}. Their results suggest that heuristics that exhibit strong correlation with the distance-to-go are less likely to produce large local minima. And large local minima are thought to often dominate the runtime of greedy planners~\cite{hoffmann2005ignoring,hoffmann2011ignoring}. They later use the Kendall rank correlation coefficient ($\tau$) to select a pattern database for some of these domains~\cite{wilt2015building}. Their use of $\tau$ as a heuristic quality metric differs from our own use because they score $\tau$ using sampled states near the goal while we score $\tau$ by ranking the states on a plan.

\section{Planning domains and training data}

Our goal is to learn a heuristic that will
improve the coverage, or the number of problems solved, for greedy
forward-search planning on 
very large satisficing planning problems. Secondary goals are to decrease the
resulting plan length and 
time to solve these problems. The search control of our planners is
greedy best first search (GBFS) with alternating, dual open
lists~\cite{richter2009preferred}. The preferred operators in the
second open list are computed by the base heuristic which, as we will
later see, is used to generate our learning
features~\cite{HoffmannN01}. We use the lazy variant
of greedy best first search which defers heuristic evaluation of
successors.
We consider STRIPS planning problems~\cite{Fikes71} with unit costs, and without axioms or conditional effects, but our techniques can be straightforwardly generalized to handle them.

\begin{defn}[Planning Domain]
A planning domain ${\cal D} = \langle {\cal P}, {\cal A} \rangle$ consists of a set of predicate schemas ${\cal P}$ and a set of action schemas ${\cal A}$. Each action schemas contains a set of precondition predicates and effect predicates. A predicate schema or action schema can be instantiated by assigning objects to its arguments.
\end{defn}

\begin{defn}[Planning Problem]
A planing problem $\Pi = \langle {\cal D}, O, s_0, g, \rangle$ is given by a domain ${\cal D}$, a set of objects $O$, an initial state $s_0$, and a goal partial-state $g$. The initial state $s_0$ is fully specified by a set of predicates. The goal partial-state $g$ is only partially specified by its set of predicates.
\end{defn}


The overall approach will be, for each {\em planning domain}, to train
a learning algorithm on several {\em planning problem} instances, and
then to use the learned heuristic to improve planning performance on
additional planning problems from that same domain.  Note
that the new problem instances use the same predicate and
action schemas, but may have different universes of objects, initial states,
and goal states.

In order to learn a heuristic for a particular domain, we must first
gather training examples from a set of existing training problems
within the domain~\cite{jimenez2012review}. Suppose that we have a
distribution over problems for a domain ${\cal D}$, which
will be used to generate testing problems. We will sample a set of
training problems $\{\Pi^1, ..., \Pi^n\}$ from ${\cal D}$. From each
problem $\Pi^i$, we generate a set of training examples in which the
$j$th training example is the pair $\langle x^i_j, y^i_j \rangle$
where $x^i_j = \langle s^i_j, \Pi^i \rangle$ is the input composed of
a state $s^i_j$ and the problem $\Pi^i$.  
Let $y^i_j$ be the length of a plan from $s^i_j$ to $g^i$. Ideally,
$y^i_j$ would be the length of the shortest plan, but because
obtaining optimal plans is intractable for the problems we consider,
we construct approximately optimal plans and use their lengths as
the $y$ values in the training data. 

We use the set of states on a single high-quality plan from
the initial state to the goal state as training
examples. 
Unfortunately, we have observed that using low-quality plans,
which are more easily found, can be dangerous, as it introduces large
amounts of noise into the training data. This noise can produce
conflicting observations of $y^i_j$ for similar $x^i_j$, which can
prevent the learner from identifying any meaningful predictive
structure. 
Reducing at least this kind of local noise is important for the
learning process even if the global plan is still suboptimal. 
Thus, we post-process each candidate plan using two local search methods: action
elimination and plan neighborhood graph
search~\cite{nakhost2010action}. 

In separate experiments, we attempted learning
a heuristic by instead using a sampled set of successors on
these plans as training examples. However, we found that the inclusion of these states 
slightly worsened the resulting performance of the learners. Our hypothesis is that the 
inclusion of successor states improves local accuracy at the expense of global accuracy.
Because the runtime of greedy search methods is often dominated by the time to escape
the largest local minima~\cite{hoffmann2005ignoring,hoffmann2011ignoring,wilt2012does,wilt2015building}, it is a worthwhile tradeoff to reduce the size of large local minima at the cost of increasing the size of small local minima. 



\section{Feature Representation}

The majority of machine learning methods assume that the inputs are represented as points in a vector space.  In our case, the inputs $x^i_j$ are a pair of a state and a planning problem, each of which is a complex structured symbolic object.  So, we need to define a feature-mapping function $\phi$ that maps an $x$ value into a vector of numeric feature values.  This can also be done implicitly by defining a kernel, we restrict our attention to finite-dimensional $\phi$ that are straightforwardly computable.

The objective in designing a feature mapping is to arrange for
examples that are close in feature space to have similar output
values.  Thus, we want to reveal the structural aspects of an input
value that encode important similarities to other input values. This
can be particularly challenging in learning for planning:  while
problems within the same domain share the same schemas for predicates
and actions, the set of objects can be arbitrarily different. For
example, a feature representation with a feature for each predicate
instance present in $s^i_j$ or $g^i$ will perform poorly on new
problems, which may not share any predicate instances
with the problems used to create the feature representation. 


Yoon et al. used information from the FF heuristic to construct
additional features from the resulting relaxed
plan~\cite{yoon2006learning,yoon2008learning}. The relaxed plan
compresses the large set of possible actions into a small plan of
actions that are likely to be relevant to achieving the goal. 
Many modern heuristics either explicitly or implicitly generate approximate plans, similar to FF's relaxed plan, that can be represented as directed acyclic graphs (DAG) where each action is a vertex, and directed edges indicate that the outgoing action is supported by the incoming action. We provide feature mappings that are applicable to any heuristic that gives rise to such a DAG, but in this paper, we focus on the FF~\cite{HoffmannN01} and CEA~\cite{helmert2008unifying} heuristics. Our method can be extended to include additional features for example derived from landmark heuristics or domain-dependent heuristics, although we do not consider these extensions here.





We can now view our training inputs as $x^i_j = \langle s^i_j, g^i, \pi^{ij}_h \rangle$ where $\pi^{ij}_h$ is the DAG generated by heuristic $h$ for state $s^i_j$ and goal $g^i$. The computation time of each feature affects the performance of the resulting planner in a complex way: the feature representation is computed for every state encountered in the search, but good features will make the heuristic more effective, causing fewer states to be encountered.

\subsection{Single Actions}

The first feature representation serves primarily as a baseline. Each
feature is the number of instances of a particular action schema in
the DAG $\pi^{ij}_h$. The number of features is the number of action
schemas $|{\cal A}|$ in the domain and thus around five for many
domains. This feature representation is simple and therefore limited
in its expressiveness,  but it can be easily computed in time $O(|\pi^{ij}_h|)$ and is unlikely to overfit. If we are learning a linear function of $\phi(x)$, then the weights can be seen as adjustments to the predictions made by the DAG of how many instances of each action are required.  So, for instance, in a domain that requires a robot to do a "move" action every time it "picks" an object, but where the delete relaxation only includes one "move" action, this representation would allow learning a weight of two on pick actions, effectively predicting the necessity of extra action instances.

\subsection{Pairwise Actions}

The second feature representation creates features for pairs of actions, encoding both their 
intersecting preconditions and effects as well as their temporal ordering in the approximate plan.
First, we solve the all-pairs shortest paths problem on $\pi^{ij}_h$ by running a BFS from each action vertex.
Then, consider each pair of actions $a_1 \to a_2$ where $a_2$ descends from $a_1$, as indicated by having
a finite, positive distance from $a_1$ to $a_2$ in the all-pairs shortest paths solution. This indicates 
$a_2$ must come after $a_1$ on all topological sorts of the DAG; i.e., $\pi^{ij}_h$ contains the implicit partial ordering $a_1 \prec a_2$.
Moreover, if there is an edge $(a_1, a_2)$ in $\pi^{ij}_h$, then $a_1 \prec a_2$ is an explicit partial ordering because $a_1$
directly supports $a_2$.

For every pair of action schemas $(A_1, A_2)$, we include two
features, counting the number of times it happens that, for an
instance $a_1$ of $A_1$ and instance $a_2$ of $A_2$, 
\begin{enumerate}[noitemsep,nolistsep]
\item $a_1 \prec a_2$, $\proc{eff}(a_1) \cap \proc{pre}(a_2) \neq \emptyset$
\item $a_2 \succ a_1$, $\proc{eff}(a_2) \cap \proc{pre}(a_1) \neq \emptyset$
\end{enumerate}
The current state and goal partial-state are included as dummy actions with only effects or preconditions respectively.

This feature representation is able to capture information about the temporal spread of actions in the DAG:
for example, whether the DAG is composed of many short parallel sequences of actions or a single long sequence.
Additionally, the inclusion of the preconditions and effects that overlap encodes interactions that are not often directly captured in the base heuristic. For example, FF and CEA make predicate independence approximations, which can result in overestimating the distance-to-go. The learner can automatically correct for these estimations if it learns that a single sequence can be used to achieve multiple predicates simultaneously.

In contrast to the single-action feature representation, the
computation of the pairwise representation takes $O(|\pi^{ij}_h|^2)$
in the worst
case. However, the DAG frequently is composed of almost disjoint
subplans, so in practice, the number of pairs considered is fewer than
$|\pi^{ij}_h| \choose 2$. Additionally, this tradeoff is still
advantageous if the learner is able to produce a much better
heuristic. Finally, for both the single and pairwise feature
representations, we add three additional features corresponding to the
original heuristic value, the number of layers present in the DAG, and
the number of unsatisfied goals.

\section{Models for heuristic learning}

We consider two different framings of the problem of learning a
heuristic function $f$.  In the first, the goal is to ensure that the
$f(x)$ values are an accurate estimate of the distance-to-go in the
planning state and problem encoded by $x = (s, \Pi)$.  In the second, the goal is
to ensure that the $f(x)$ values accurately rank the distance-to-go
for different states $s$ within the same planning problem 
$\Pi$, but do not necessarily reflect that actual distance-to-go values.

These different framings of the problem lead to different loss
functions to be optimized by the learner and to different optimization
algorithms.  Because our learning algorithms cannot optimize for
search performance directly, the loss function serves as a proxy for
the search performance. A good loss function will be highly correlated
with performance of learned heuristics. We restrict ourselves to linear models that learn a weight vector $w$,
and make a prediction $f(x) = \phi(x)^T w$. 

\subsection{Heuristic value regression}

Because learning a heuristic is, at face value, a regression problem, a natural loss function is the root mean squared error (RMSE). A model with a low RMSE produces predictions close to the actual distance-to-go. Because each training problem $\Pi^i$ may produce a different number of examples $m_i$, we use the average RMSE over all problems. This ensures that we do not assign more weight to problems with more examples. If $f$ is a prediction function mapping a vector to the reals, then:
\begin{equation*}
\text{RMSE} = \frac{1}{n} \sum_{i=1}^n \sqrt{\frac{1}{m_i} \sum_{j=1}^{m_i} (f(x^i_j) - y^i_j)^2}. 
\end{equation*}
The first learning technique we applied is ridge regression
(RR)~\cite{hoerl1970ridge}. This serves as a baseline to compare to
the results of Yoon et al.~\cite{yoon2008learning}. Ridge regression
is a regularized version of Ordinary Least Squares Regression
(OLS). The regularization trades off optimizing the squared error
against preferring low magnitude $w$ using a parameter $\lambda$. This
results in the following optimization problem. Leting $\phi(X)$ be the
design matrix of concatenated features $\phi(x^i_j)$ and $Y$ be the
vector concatenation of $y^i_j$ for all $i, j$, 
we wish to find
\begin{equation*}
\min_w ||\phi(X) w - Y||^2 + \lambda ||w||^2 \;\;.
\end{equation*}
This technique is advantageous because it can be quickly solved in
closed form for reasonably sized $\phi(X)$, yielding the weight vector
\begin{equation*}
w = (\phi(X)^T\phi(X) + \lambda I)^{-1}Y\;\;.
\end{equation*}
Optimizing
RMSE directly, with no penalty $\lambda$, will yield a weight vector
that performs well on the training data but might not generalize well
to previously unseen problems. Increasing $\lambda$ forces the
magnitude of $w$ to be smaller, which prevents the resulting $f$ from
"overfitting" the training data and therefore not generalizing well to
new examples. This is especially important in our application as we
are trying learn a heuristic that generalizes across the full
state-space from only a few representative plans.  

We select an appropriate value of $\lambda$ by performing domain-wise
leave-one-out cross validation (LOOCV): For different possible values
of $\lambda$, and in a domain with $n$ training problem, we
train on data from $n-1$ training problems and evaluate the resulting
heuristic on the remaining problem according to the RMSE loss
function, and average the scores from holding out each problem
instance. We select the $\lambda$ value for which the LOOCV RMSE is
minimized over a logarithmic scale.


\subsection{Learning to Rank}

The RMSE, however, is not the most appropriate metric for our learning
application. We are learning a heuristic for greedy search, which uses
the heuristic solely to determine open list priority. The value of the
heuristic per se does not govern the search performance which depends
most directly on the ordering on states induced by the heuristic. In
this context, any monotonically increasing function of a heuristic
results in the same ranking and performance. A heuristic may have
arbitrarily bad RMSE despite performing well.

For these reasons, we consider the Kendall rank correlation
coefficient ($\tau$), a nonparametric ranking statistic, as a loss
function. It represents the normalized difference between the number
of correct rankings and incorrect rankings for each of the ranking
pairs. As with the RMSE, we compute the
average $\tau$ across each problem. The separation of problems is even
more important here. Our $\tau$ only scores rankings between
examples from the same problem as examples from separate 
problems are never encountered together in the same search.
This provides a major source of leverage over an ordinary regression
framework. Heuristics are not penalized for producing inconsistent
distances-to-go values across multiple problems, allowing them to provide
more effort to improve the per-problem rankings.

Let $s(i; j, k)$ score the concordance or discordance of the ranking function $f$ for examples $\langle x^i_j, y^i_j \rangle$ and $\langle x^i_k, y^i_k \rangle$ from the same problem $\Pi^i$:
\begin{equation*}
s(i; j, k) = \begin{cases} 
+1 &\text{sgn}(f(x^i_k) - f(x^i_j)) = \text{sgn}(y^i_k - y^i_j) \\ 
-1 &\text{sgn}(f(x^i_k) - f(x^i_j)) = -\text{sgn}(y^i_k - y^i_j) \\
0 &f(x^i_k) - f(x^i_j) = 0
\end{cases}.
\end{equation*}
Then the Kendall rank correlation coefficient is specified by 
\begin{equation*}
\tau = \frac{1}{n} \sum_{i=1}^n \frac{2}{m_i(m_i - 1)}\sum_{j=1}^{m_i} \sum_{k=j+1}^{m_i} s(i; j, k)\;\;.
\end{equation*}
Note that each $y^i_j$ is unique per problem $\Pi^i$ because our examples come from a single trajectory. Observe that $\tau \in [-1, 1]$; values close to one indicate the ranking induced by the heuristic $f$ has strong positive correlation to the true ranking of states as given by the actual labels. Conversely, values close to negative one indicate strong negative correlation. 

If our loss function is $\tau$, it is more effective to optimize $\tau$ directly in the learning process. To this end, we use Rank Support Vector Machines (RankSVM)~\cite{joachims2002optimizing}. RankSVMs are variants of SVMs which penalize the number of incorrectly ranked training examples. Like SVMs, RankSVMs also have a parameter $C$ used to provide regularization. Additionally, their formulation uses the hinge loss function to make the learning problem convex. Thus, a RankSVM finds the $w$ vector that optimizes a convex relaxation of $\tau$. Our formulation of the RankSVM additionally takes into account the fact that we only wish to rank training examples from the same problem. Our formulation is the following:
\begin{equation*}
\begin{aligned}
& \underset{w}{\text{min}}
& & ||w||^2 + C\sum_{i=1}^{n}\sum_{j=1}^{m_i}\sum_{k=j+1}^{m_i}\xi_{ijk} \\
& \text{s.t.} & &  \phi(x^i_j)^T w \geq \phi(x^i_k)^T w + 1 - \xi_{ijk}, \; \forall y^i_j \geq y^i_k, \; \forall i \\
& & &  \xi_{ijk} \geq 0, \; \forall i, j, k \;\;.\\
\end{aligned}
\end{equation*}
The first constraint can also be rewritten to look similar to the original SVM formulation. In this form, the RankSVM can be viewed as classifying if $x^i_j, x^i_k$ are properly ranked.
\begin{equation*}
(\phi(x^i_j) - \phi(x^i_k))^T w \geq 1 - \xi_{ijk}, \; \forall y^i_j \geq y^i_k, \; \forall i
\end{equation*}
Notice that number of constraints and slack variables, corresponding
to the number of rankings, grows quadratically in the size of each
problem. This makes training the RankSVM more computationally expensive than RRs or SVMs. However, there are efficient methods for training these, and other SVMs, when considering just the linear, primal form of the problem~\cite{joachims2006training,franc2009optimized}. 
It is important to note that we generate a number of constraints that is quadratic only in the length of any given training plan, and do not attempt to rank all the actions of all the training plans jointly; this allows us to increase the number of training example plans without dramatically increasing the size of the optimization problem.

An additional advantage of RankSVM is that it supports the inclusion
of the non-negativity constraint $w \geq 0$ which provide 
additional regularization. Because each feature
represent a count of actions or action pairs, the values are
always non-negative, as are the target values. 
We generally expect that DAGs with
a large number of actions indicate that the state is far from the
goal. The non-negativity constraint allows us to incorporate this
prior knowledge in the model, which can sometimes improve the
generalization of the learned heuristic.
As in RR, we select $C$ using a line search over a logarithmic scale,
to maximize a cross-validated estimate of $\tau$.  As a practical
note, we start with an over-regularized model where $C \approx 0$ and
increase $C$ until reaching a local minimum because SVMs are trained much
more efficiently for small $C$.  

\section{Results}

\begin{table*}[t!]
\centering
\small
\begin{tabular}{|l|r|r|r|r|r|r|r|r|r|r|}
\hline
\multicolumn{11}{|c|}{{\it elevators} (35)}                                                                                                                                      \\ \hline
\textbf{Method} & \textbf{Cov.} & \textbf{Len.} & \textbf{Run T.} & \textbf{Exp.} &  & \textbf{RMSE} & \textbf{$\tau$} & \textbf{$\lambda$/$C$} & \textbf{Feat.} & \textbf{Train T.} \\ \cline{1-5} \cline{7-11} 
FF Original     & 14 & 318 & 196 & 17833                &  &               34.370 & 0.9912 & N/A & N/A & N/A                  \\ \cline{1-5} \cline{7-11} 
FF RR Single    & 22 & 546 & 504 & 34970               &  &               4.091 & 0.9948 & 100 & 9/9 & 3.133                   \\ \cline{1-5} \cline{7-11} 
FF RR Pair      & 15 & 561 & 308 & 20985              &  &               3.789 & 0.9971 & 1000 & 53/53 & 11.686                   \\ \cline{1-5} \cline{7-11} 
FF RSVM Single  & 34 & 375 & 403 & 23765               &  &               79.867 & 0.9967 & 0.1 & 9/9 & 55.681                   \\ \cline{1-5} \cline{7-11} 
FF RSVM Pair   & 34 & 631 & 123 & 7083               &  &               418.828 & 0.9996 & 1 & 53/53 & 140.786                  \\ \cline{1-5} \cline{7-11} 
\textbf{FF NN RSVM Pair} & \textbf{35} & 655 & 61 & 10709               &  &               46.296 & 0.9992 & 1 & 51/53 & 125.702                   \\ \cline{1-5} \cline{7-11}
\textbf{CEA Original}     & \textbf{35} & 397 & 163 & 4504                &  &               21.494 & 0.9973 & N/A & N/A & N/A                    \\ \hline
\multicolumn{11}{|c|}{{\it transport} (35)}                                                                                                                                    \\ \hline
FF Original     & 5 & 588 & 470 & 18103                &  &                 126.193 & 0.8460 & N/A & N/A & N/A                \\ \cline{1-5} \cline{7-11} 
FF RR Single    & 0 & None & None & None             &  &              31.518 & 0.9303 & 100 & 6/6 & 3.569                   \\ \cline{1-5} \cline{7-11} 
FF RR Pair      & 4 & 529 & 560 & 27866             &  &               27.570 & 0.9392 & 10000 & 32/32 & 11.028                  \\ \cline{1-5} \cline{7-11} 
FF RSVM Single  & 21 & 1154 & 650 & 29452               &  &               149.003 & 0.9720 & 0.1 & 6/6 & 106.901                   \\ \cline{1-5} \cline{7-11} 
FF RSVM Pair    & 20 & 587 & 178 & 8896              &  &               162.141 & 0.9797 & 0.001 & 32/32 & 117.808                \\ \cline{1-5} \cline{7-11} 
\textbf{FF NN RSVM Pair} & \textbf{31} & 663 & 206 & 7803               &  &              141.273 & 0.9798 & 0.01 & 17/32 & 287.586                   \\ \cline{1-5} \cline{7-11}
CEA Original     & 9 & 448 & 542 & 9064                &  &               57.819 & 0.9314 & N/A & N/A & N/A                   \\ \cline{1-5} \cline{7-11} 
CEA RR Single    & 11 & 493 & 436 & 6921              &  &             33.032 & 0.9420 & 10000 & 6/6 & 4.536                   \\ \cline{1-5} \cline{7-11} 
CEA RR Pair      & 2 & 609 & 1602 & 40327             &  &              30.731 & 0.9318 & 100 & 45/45 & 15.716                  \\ \cline{1-5} \cline{7-11} 
CEA RSVM Single  & 18 & 722 & 588 & 11334               &  &              130.653 & 0.9748 & 0.1 & 6/6 & 158.523                   \\ \cline{1-5} \cline{7-11} 
\textbf{CEA RSVM Pair}    & \textbf{31} & 650 & 225 & 3526             &  &               159.139 & 0.9804 & 0.0001 & 45/45 & 244.164                  \\ \cline{1-5} \cline{7-11} 
CEA NN RSVM Pair & 29 & 696 & 277 & 9006            &  &              191.064 & 0.9795 & 0.0001 & 29/45 & 528.665                   \\ \hline
\multicolumn{11}{|c|}{{\it parking} (10)}                                                                                                                                      \\ \hline
FF Original     & 0 & None & None & None                &  &                 6.101 & 0.9525 & N/A & N/A & N/A               \\ \cline{1-5} \cline{7-11} 
FF RR Single    & 0 & None & None & None              &  &              4.571 & 0.9648 & 100 & 7/7 & 0.201                   \\ \cline{1-5} \cline{7-11} 
FF RR Pair      & 2 & 156 & 1419 & 33896             &  &               4.285 & 0.9757 & 100 & 40/40 & 0.570                \\ \cline{1-5} \cline{7-11} 
FF RSVM Single  & 0 & None & None & None               &  &               10.468 & 0.9745 & 0.01 & 7/7 & 8.423                   \\ \cline{1-5} \cline{7-11} 
FF RSVM Pair    & 8 & 185 & 208 & 2852              &  &               18.262 & 0.9918 & 0.1 & 40/40 & 7.030                \\ \cline{1-5} \cline{7-11} 
FF NN RSVM Pair & 6 & 183 & 358 & 5891               &  &            143.063 & 0.9941 & 10 & 26/40 & 7.119                  \\ \cline{1-5} \cline{7-11}
CEA Original     & 0 & None & None & None                &  &               15.885 & 0.9628 & N/A & N/A & N/A                  \\ \cline{1-5} \cline{7-11} 
CEA RR Single    & 0 & None & None & None              &  &               4.667 & 0.9669 & 0.01 & 7/7 & 0.277                   \\ \cline{1-5} \cline{7-11} 
CEA RR Pair      & 1 & 280 & 1230 & 48180             &  &               4.448 & 0.9660 & 10 & 47/47 & 0.738                \\ \cline{1-5} \cline{7-11} 
CEA RSVM Single  & 0 & None & None & None                &  &              7.950 & 0.9757 & 0.1 & 7/7 & 10.830                  \\ \cline{1-5} \cline{7-11} 
\textbf{CEA RSVM Pair}    & \textbf{10} & 272 & 81 & 2147             &  &  45.823 & 0.9918 & 1 & 47/47 & 10.237   \\ \cline{1-5} \cline{7-11} 
\textbf{CEA NN RSVM Pair} & \textbf{10} & 260 & 70 & 1690               &  &  140.297 & 0.9938 & 10 & 27/47 & 9.179                   \\ \hline
\multicolumn{11}{|c|}{{\it no-mystery} (10)}                                                                                                                                      \\ \hline
\textbf{FF Original}     & \textbf{4} & 31 & 583 & 5658745                &  &                 3.462 & 0.9841 & N/A & N/A & N/A                \\ \cline{1-5} \cline{7-11} 
\textbf{FF RR Single}    & \textbf{4} & 30 & 1004 & 8385159              &  &               1.662 & 0.9861 & 100 & 6/6 & 0.085                   \\ \cline{1-5} \cline{7-11} 
FF RR Pair      & 2 & 31 & 700 & 3898861             &  &               1.622 & 0.9902 & 1000 & 21/21 & 0.193                  \\ \cline{1-5} \cline{7-11} 
FF RSVM Single  & 1 & 26 & 1411 & 16201215               &  &               21.069 & 0.9871 & 100 & 6/6 & 0.712                   \\ \cline{1-5} \cline{7-11} 
FF RSVM Pair    & 2 & 28 & 892 & 6894959              &  &              39.350 & 0.9968 & 1 & 21/21 & 0.914                 \\ \cline{1-5} \cline{7-11} 
FF NN RSVM Pair & 1 & 29 & 1049 & 7973003               &  &              80.588 & 0.9972 & 10 & 17/21 & 1.024                 \\ \cline{1-5} \cline{7-11}
CEA Original     & 3 & 30 & 73 & 107773               &  &               16.851 & 0.9579 & N/A & N/A & N/A                  \\ \cline{1-5} \cline{7-11} 
CEA RR Single    & 2 & 28 & 9 & 26319              &  &               1.824 & 0.9890 & 100 & 6/6 & 0.069                  \\ \cline{1-5} \cline{7-11} 
CEA RR Pair      & 3 & 32 & 104 & 169434            &  &               1.717 & 0.9892 & 1000 & 32/32 & 0.342                  \\ \cline{1-5} \cline{7-11} 
CEA RSVM Single  & 2 & 28 & 12 & 33559               &  &               36.457 & 0.9916 & 1 & 6/6 & 1.283                  \\ \cline{1-5} \cline{7-11} 
CEA RSVM Pair    & 3 & 32 & 34 & 46501              &  &               6.358 & 0.9964 & 0.01 & 32/32 & 4.023                 \\ \cline{1-5} \cline{7-11} 
CEA NN RSVM Pair & 3 & 31 & 190 & 264225              &  &              55.141 & 0.9970 & 1 & 16/32 & 62.608                  \\ \hline
\end{tabular}
\caption{Results from the {\it elevators}, {\it transport}, {\it parking}, and {\it no-mystery} IPC Learning Track 2014 problems.}
\label{no-mystery}
\end{table*}

We implemented our planners using the FastDownward
framework~\cite{helmert2006fast}.
\footnote{Because heuristic values
  are required to be integers in this framework, we scale up and then
  round predicted heuristic values, in order to capture more of the
  precision in the values. 
  Recall that scaling
  will not alter the planner's performance because arbitrary
  nonnegative, affine transformations to $f(x)$ will not affect the
  resulting ranking in greedy search.
 }
Each planning problem is compiled to
a representation similar to SAS+~\cite{backstrom1995complexity} using
the FastDownward preprocessor. However, the predicates that represent
each SAS+ (variable, value) pair are still stored so, actions and
states can be mapped back to their prior form. We used the {\tt dlib}
C++ machine learning library to implement the learning
algorithms~\cite{king2009dlib}.  

We experimented on four domains from the 2014 IPC learning
track~\cite{ipc2014}: {\it elevators}, {\it transport}, {\it parking},
and {\it no-mystery}. 
For each domain, we
constructed a set of unique examples with the competition problem
generators by sampling parameters that cover competition parameter
space. We use a variant of the 2014 FastDownward Stone Soup
portfolio~\cite{helmert2011fast} planner, with a large timeout and memory limit, to generate training example plans. 
We trained on at most 10 examples randomly selected from the
set of problems our training portfolio planner was able to solve, and
then tested on the remaining problems. 


For each experiment, we report the following
values:\footnote{We use arithmetic mean for
  plan length and geometric means for planning time and number of
  expansions, and report these statistics only for solved instances;
  RMSE and $\tau$ values are cross-validation estimates.}
{\bf Cov:} {\em coverage}, or total number of problems solved; 
{\bf Len:} mean plan length; 
{\bf Run T:} mean planning time in seconds; 
{\bf Exp:} mean number of expansions; 
{\bf RMSE:} RMSE of learned heuristic,
$\tau$: Kendall rank correlation coefficient of learned
  heuristic; 
$\lambda$/$C$: regularization parameter value 
  ($\lambda$ for RR and $C$ for RankSVM);
{\bf Feat:} number of nonzero weights learned relative to the
  total number of features; and 
{\bf Train T:} runtime to train the heuristic learner in seconds.

Each planner was run on a single 2.5 GHz processor for 30 minutes
with 5 GB of memory. We only include the results of the original CEA
heuristic on {\it elevators}, as the default heuristic was able to
solve each problem and the heuristics learned using CEA all performed
similarly. 

The heuristics learned by RankSVM are able to solve more problems than
those learned using ridge regression. Within a domain, $\tau$ seems to be
positively correlated with the number of problems solved while the
RMSE does not. The pairwise-action
features outperform the single-action features in
RankSVM, making it worthwhile to incur a larger heuristic
evaluation time for improved heuristic strength. The CEA learned heuristics performed
slightly better than the FF learned heuristics.

On {\it transport} and {\it parking}, the training portfolio planner was 
only able to solve the smallest problems within the parameter space. Thus, our RankSVM learners
demonstrate the ability to learn from smaller problems and perform well on larger problems.
In separate experiments, we observed that both artificially over-regularized and under-regularized learners 
performed poorly indicating that selection of the regularization parameter is important to the learning process.

The learned heuristics perform slightly worse than
the standard heuristics on {\it no-mystery} despite having almost perfect $\tau$ values.
In separate experiments using eager best-first search, the learned heuristics perform slightly
better on {\it no-mystery}, but the improvement is not significant.
This domain is known to be challenging for heuristics because it contains a large number of dead-ends.
We observed that $\tau$ does not seem sufficient for understanding heuristic performance on domains with harmful dead-ends. 
Our hypothesis is that failing to recognize a dead-end is often more harmful than incorrectly ranking nearby states and should be handled separately from learning a heuristic. A topic for future work is to combine our learned heuristics with learned dead-end detectors.

Inclusion of the non-negativity constraint (NN) on {\it transport} significantly improved the coverage of the FF learned heuristic over the normal RankSVM formulation. We believe that this constraint can sometimes improve generalization in domains with a large variance in size or specification. For example, the {\it transport} generator samples problems involving either two or three cities leading to a bimodal distribution of problems. 

Finally, we tested two learned heuristics on 
the five evaluation problems per domain chosen in the IPC 2014 learning track.
Both the FF RSVM Pair heuristic and the CEA RSVM Pair heuristic solved 
all 5/5 problems in {\it elevators}, {\it transport}, and {\it parking} but only 1/5 problems in {\it no-mystery}.

\section{Conclusion} 

Our results indicate that, for greedy search, learning a heuristic is
best viewed as a ranking problem. The Kendall rank correlation
coefficient $\tau$ is a better indicator of a heuristic's quality than
the RMSE, and it is effectively optimized using the RankSVM learning
algorithm. 
Pairwise-action features outperformed simpler
features. 
Further work involves combining features from several heuristics, 
learning complementary search control using our features, and
incorporating the learned heuristics in planning portfolios. 

\section*{Acknowledgments} 

We gratefully acknowledge support from NSF grants 1420927 and 1523767, from ONR grant N00014-14-1-0486, and from ARO grant W911NF1410433.  Any opinions, findings, and conclusions or recommendations expressed in this material are those of the authors and do not necessarily reflect the views of our sponsors.


\newpage
\bibliographystyle{named}
\bibliography{references}

\end{document}